\documentclass[runningheads]{llncs}
\usepackage[T1]{fontenc}
\usepackage{graphicx,verbatim}
\usepackage{amsmath}
\usepackage{amssymb}
\usepackage{booktabs}
\usepackage{multirow}
\usepackage{multicol}
\usepackage{tabularx}
\usepackage{utfsym}
\usepackage{array}
\begin{document}
\title{Large Language Model Aided Birt-Hogg-Dubé Syndrome Diagnosis with Multimodal Retrieval-Augmented Generation}
\titlerunning{BHD Diagnosis with Multimodal Retrieval-Augmented Generation}
%

\author{Haoqing Li, Jun Shi, Xianmeng Chen, Qiwei Jia, Rui Wang, Wei Wei, Hong An, Xiaowen Hu}  
\authorrunning{Anonymized Author et al.}
\institute{School of Computer Science and Technology \\
    \email{li\_haoqing@mail.ustc.edu.cn}, \\
    Department of Pulmonary and Critical Care Medicine; Center for Diagnosis and Management of Rare Diseases, the First Affiliated Hospital of USTC, Division of Life Sciences and Medicine, USTC, \\
    WanNan Medical College
}

\maketitle              
\begin{abstract}
Deep learning methods face dual challenges of limited clinical samples and low inter-class differentiation among Diffuse Cystic Lung Diseases (DCLDs) in advancing Birt-Hogg-Dubé syndrome (BHD) diagnosis via Computed Tomography (CT) imaging. While Multimodal Large Language Models (MLLMs) demonstrate diagnostic potential for such rare diseases, the absence of domain-specific knowledge and referable radiological features intensify hallucination risks. To address this problem, we propose BHD-RAG, a multimodal retrieval-augmented generation framework that integrates DCLD-specific expertise and clinical precedents with MLLMs to improve BHD diagnostic accuracy. BHD-RAG employs: (1) a specialized agent generating imaging manifestation descriptions of CT images to construct a multimodal corpus of DCLDs cases. (2) a cosine similarity-based retriever pinpointing relevant image-description pairs for query images, and (3) an MLLM synthesizing retrieved evidence with imaging data for diagnosis. BHD-RAG is validated on the dataset involving four types of DCLDs, achieving superior accuracy and generating evidence-based descriptions closely aligned with expert insights.


\end{abstract}
\section{Introduction}

\begin{figure}[t]
	\includegraphics[width=\textwidth]{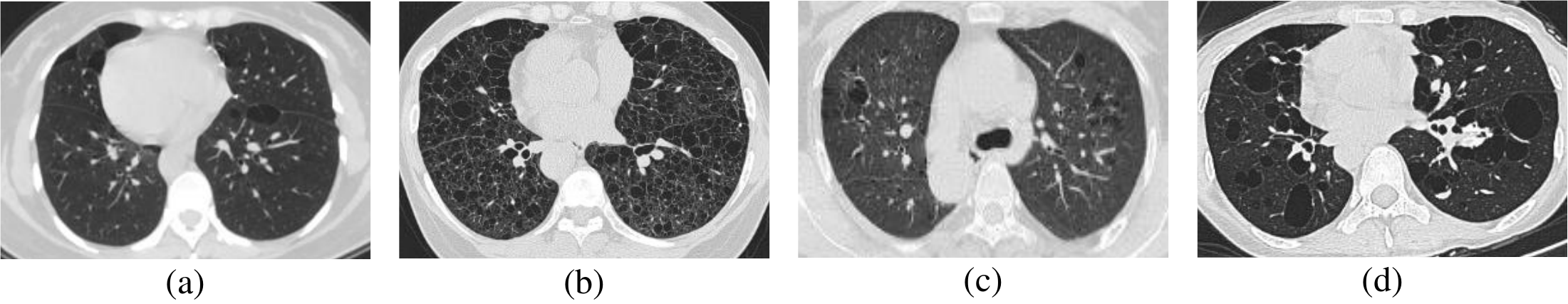}
	\caption{Representative CT imaging examples of four challenging-to-differentiate DCLDs: (a) BHD, characterized by cysts of variable size, elliptical or flattened, predominantly located in the lower lungs and subpleural regions; (b) LAM, featuring round and of similar size cysts; (c) PLCH, predominantly located in the lower lungs and subpleural regions; (d) LIP, the cysts vary in size, predominantly in the bilateral lung bases, and follow a perivascular pattern.}
	\label{fig1}
\end{figure}

Birt-Hogg-Dubé syndrome (BHD) is an autosomal dominant disorder characterised clinically by skin fibrofolliculomas, pulmonary cysts, spontaneous pneumothorax, and renal cancer \cite{1,2}. The BHD diagnosis relies on identifying cystic lesions and pneumothorax in chest Computed Tomography (CT) scans. However, it remains challenging to distinguish BHD from other Diffuse Cystic Lung Diseases (DCLDs) with similar radiological features, such as numerous diffusely distributed, thin-walled, round, or irregular pulmonary cystic lesions. These DCLDs mainly include Lymphangioleiomyomatosis (LAM), Pulmonary Langerhans Cell Histiocytosis (PLCH), and Lymphocytic Interstitial Pneumonia (LIP) \cite{2,3}. Therefore, accurate differentiation between BHD and other DCLDs is critical for precise diagnosis and appropriate clinical management.

In recent years, deep learning methods have demonstrated remarkable performance in challenging diagnostic tasks, comparable to human experts \cite{4,5,6}. However, existing task-specific methods often rely on  large-scale training data and do not apply to rare diseases such as BHD with limited clinical samples. Furthermore, the low inter-class differentiation in imaging manifestations among DCLDs hinder traditional discriminative methods from optimizing complex decision boundaries, limiting their classification performance for BHD.

With the In-Context Learning (ICL) paradigm \cite{26,27,28,29}, Large Language Models (LLMs) have shown promise in data-scarce medical scenarios. For example, GPT-4 outperform  domain-specific models like Med-PaLM 2 \cite{8} without requiring task-specific training. \cite{7}. This technological advancement positions Multimodal Large Language Models (MLLMs) as a feasible solution for achieving accurate diagnosis of BHD syndrome in low-sample clinical settings. However, the susceptibility of MLLMs to "hallucination", i.e. generating plausible yet incorrect outputs, is a critical limitation \cite{9}. The absence of domain-specific expertise and referable radiological features of DCLDs further intensify hallucination risks, as shown in Fig. \ref{fig5}.

Retrieval-Augmented Generation (RAG) integrates information retrieval and text generation, leveraging external corpus to enhance MLLMs in knowledge-intensive tasks \cite{10}. As a typically knowledge-intensive field, medical diagnostics can benefit from RAG, enabling MLLMs to generate evidence-based medical responses and analyze complex queries \cite{11,12}. For BHD diagnosis, RAG is capable of providing domain-specific knowledge and imaging features on BHD and other DCLDs, addressing the hallucination inherent in MLLMs. However, the rarity of DCLDs and the limited availability of publicly accessible information constrain the construction of corpus.

\begin{figure}[t]
	\includegraphics[width=\textwidth]{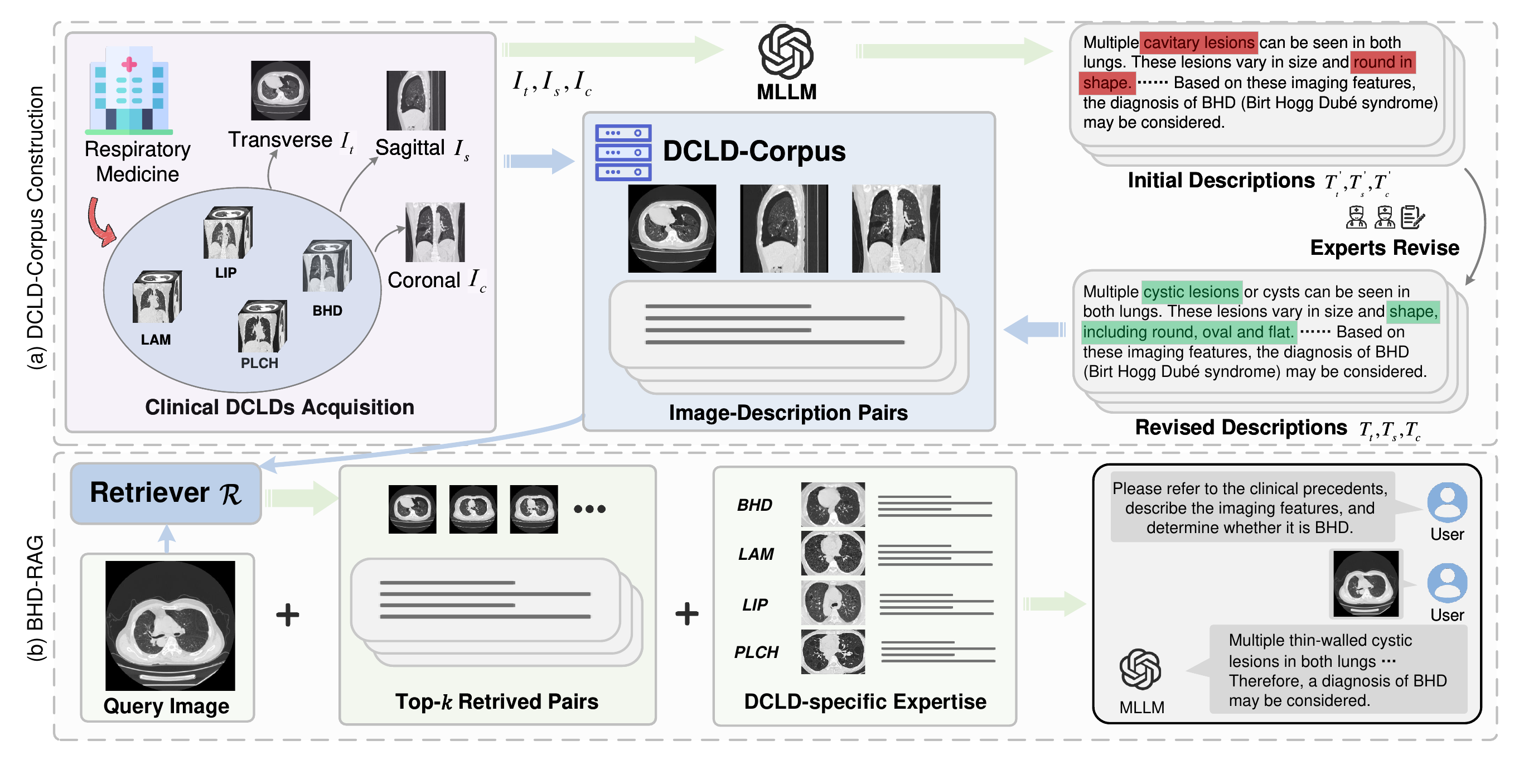}
	\caption{An overview of the proposed BHD-RAG framework.} \label{fig2}
\end{figure}

To address these challenges, we propose BHD-RAG, a multimodal retrieval-augmented generation framework that integrates DCLD-specific expertise and clinical precedents with MLLMs to enhance BHD diagnosis. As shown in Fig. \ref{fig2}, in response to the scarcity of external knowledge to construct the RAG system, we sourced clinical cases of DCLDs from the respiratory department to establish a dedicated DCLD-Corpus. MLLMs are employed to generate respiratory medicine-specific descriptions from CT slices, which are refined in collaboration with respiratory specialists to construct a corpus of image-description pairs. Moreover, to overcome the challenge of optimizing the decision boundary between BHD and other DCLDs, BHD-RAG employs a cosine-space similarity retriever to expand the angular margin and extract pertinent pairs from the corpus. The generator subsequently integrates the knowledge retrieved with the query to produce diagnostic responses. BHD-RAG is validated on the dataset encompassing four DCLDs, demonstrating superior accuracy and generating evidence-based descriptions consistent with expert assessments.

\section{Methodology}

\subsection{DCLD-Corpus Construction}
MLLMs are prone to struggling with factual accuracy when addressing challenging medical tasks. Retrieval-Augmented Generation (RAG) offers a solution by integrating external corpus to guide response generation and improve reliability. As a foundational step, it is essential to develop a DCLD-corpus for BHD diagnosis, to support the RAG system.

To develop a corpus for DCLDs, it is essential to gather extensive image-description paired diagnostic examples. We leverage the fine-grained image description capabilities of MLLMs to generate detailed respiratory medicine-specific descriptions. Firstly, We segment the CT scans into 2D slices along the coronal, sagittal, and transverse planes $\{I_t, I_s, I_c\}$, and retain abnormal slices containing prominent cystic lesions. Moreover, GPT-4-turbo \cite{16} is used to generate initial imaging manifestation descriptions $\{T'_t, T'_s, T'_c\}$ of these slices. Medical experts then refine these descriptions, correcting misclassifications to ensure the accuracy of the descriptions $\{T_t, T_s, T_c\}$ and the alignment with expert diagnostic logic, as shown in Fig. \ref{fig2} (a).

The descriptions $\{T_t, T_s, T_c\}$ generated by the experts-assisted MLLMs are paired with the corresponding images to construct the corpus:
 
\begin{equation}
	\{\mathcal{I}^i_{corpus}, \mathcal{T}^i_{corpus}\}=\{(I^i_t, T^i_t), (I^i_s, T^i_s), (I^i_c, T^i_c)\},
\end{equation}
where $i=0, 1, 2, ..., n-1$, and $n$ represents the number of pairs. Due to the limited availability of clinical samples for rare DCLDs, related data is scarce in training databases of MLLMs. Thus, maximizing the utility of our valuable clinical samples and provides the MLLM with comprehensive external knowledge of DCLDs is imperative. Moreover, the BHD diagnosis requires both an assessment of lesion morphology and the determination of its spatial location within the lungs. Multi-view imaging provides a comprehensive features of lesion morphology than single-view imaging, offering spatial evidence from multiple perspectives for MLLMs.

Furthermore, the lack of DCLD-specific knowledge in MLLMs introduces misleading contexts and extraneous information, compromising generative performance \cite{17,18,19}. To address this, DCLD-corpus is designed to emphasize distinctive DCLDs features, providing fine-grained comparative analyses to reduce domain knowledge ambiguity, as shown in Fig. \ref{fig2} (b). The constructed corpus integrates multimodal, multi-grained DCLD-specific knowledge, enabling the retriever to supply the MLLMs with the most relevant and precise information.

\subsection{Cosine-Space Similarity Retriever}

The retriever is integral to the BHD-RAG framework, tasked with extracting the minimal corpus subset essential for accurate inference by the generator \cite{20}. During the multimodal corpus retrieval, the BHD-RAG retriever selects the top-$k$ image-description pairs, which are most similar to the query CT slide. These references contain expert-curated domain-specific knowledge on DCLDs rare diseases and guide the MLLM in generating accurate responses to DCLDs-related queries. Ensuring retrieval accuracy and efficiency is crucial to this process.

\begin{figure}[t]
	\includegraphics[width=\textwidth]{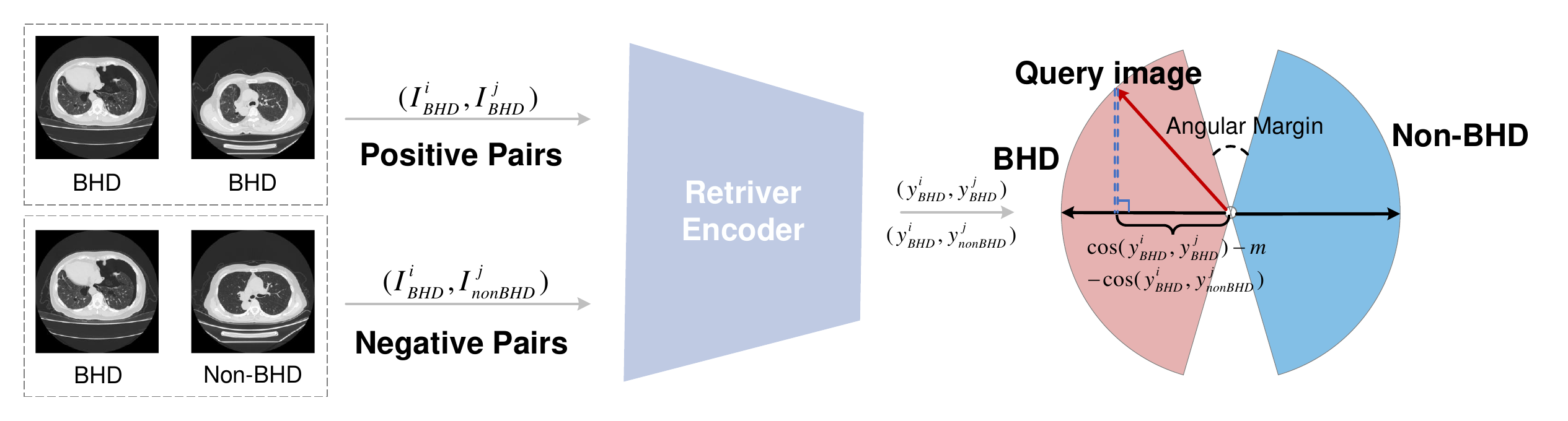}
	\caption{The proposed cosine spatial similarity measure retriever.} \label{fig3}
\end{figure}

The retriever is trained with a similarity-based metric learning paradigm \cite{13,14}, as illustrated in Fig. \ref{fig3}. Both positive and negative samples are matched within the same anatomical view. Positive pairs, e.g. $(I^i_{BHD}, I^j_{BHD})$, are constructed by randomly sampling different CT slices from the same class, while negative pairs, e.g. $(I^i_{BHD}, I^j_{nonBHD})$, are generated from different classes. $I^i_{BHD}$ represents a randomly selected $i$-th sample from the BHD slides. The output $y$ of the retriever $\mathcal{R}$ can be formulated as follows:

\begin{equation}
	(y^i_{BHD}, y^j_{nonBHD})=\mathcal{R}(I^i_{BHD}, I^j_{nonBHD}).
\end{equation}

Under the substantial overlap of imaging features across various DCLDs, conventional approaches relying on cosine similarity with Softmax struggle to achieve precise classification in cosine space, as shown in Table 1. To address this, we define the decision boundaries in the cosine space, and introduce CosFace loss to enhance the angular margin between BHD and non-BHD classes \cite{15}:

\begin{equation}
	L = \frac{1}{N} \sum_{i} -\log \frac{e^{s(\cos(y^i_{BHD}, y^j_{BHD}) - m)}}{e^{s(\cos(y^i_{BHD}, y^j_{BHD}) - m)} + e^{s \cos(y^i_{BHD}, y^j_{nonBHD})}},
\end{equation}
where $N$ denotes the total number of positive and negative pairs, $cos(.)$ represents the cosine similarity between sample logits, $s$ is the scaling factor, and $m$ is the angular margin between classes.

\subsection{Retrieval-Augmented Generation for BHD Diagnosis}
The objective is to diagnose BHD from the query image $I_q$ and generate an accurate description of its imaging characteristics, producing response $\tilde{A}$. To enhance the ability in describing and discriminating DCLDs features, the top-$k$ image-description pairs $\{\mathcal{I}_{topk}, \mathcal{T}_{topk}\}$ retrieved from the corpus are combined with expert-curated DCLD-specific expertise $\{I_{e}, T_{e}\}$. This integration strengthens the performance of the MLLMs in knowledge-intensive BHD diagnostic tasks. The output of BHD-RAG is formalized as: 

\begin{equation}
	\{\mathcal{I}_{topk}, \mathcal{T}_{topk}\} =
	\mathop{\arg\min}_{\substack{\mathcal{I}_{topk} \in \mathcal{I}_{corpus} \\ |\mathcal{I}_{topk}| = k}}
	\left\{ \cos\left(\mathcal{R}(I_q, \mathcal{I}_{corpus}^i)\right) \right\}, \, i = 0, 1, \ldots, n - 1,
\end{equation}

\begin{equation}
	\tilde{A} = f(I_q, T_q; \mathcal{M}, \mathcal{I}_{corpus}, \mathcal{T}_{corpus}) = \mathop{\arg\min}_A \mathbb{P}_{\mathcal{M}}(A \mid I_q, I_{e}, \mathcal{I}_{topk}, T_q, T_{e}, \mathcal{T}_{topk}),
\end{equation}
where $f(.)$ represents the proposed BHD-RAG, $T_q$ is the input prompt, $\mathcal{M}$ denotes the MLLM, $\tilde{A}$ and $A$ are the predicted response and the ground truth, respectively.

\section{Experiments and Results}

\subsection{Data Curation and Implementation Details}

CT scans from patients with confirmed diagnoses of DCLDs based on histopathological confirmation or in accordance with well-established professional society guidelines, were acquired from the **** Hospital. The contiguous DICOM images, with a median slice thickness of 1.25 mm (range: 1–5 mm), comprise 97 cases (50 BHD, 18 LAM, 7 PLCH, and 22 LIP).

These data are divided into train (retrieve) and test (query) set in an 8 : 2 ratio at patient-level, with slices $I_t, I_s, I_c$ containing typical DCLDs lesions identified. To ensure diversity in the corpus, adjacent key slices were spaced by at least two frames. Additionally, slices exhibiting typical lesions are prioritized to maximize diagnostic information. The numbers of typical slices for BHD, LAM, PLCH, and LIP were 753, 255, 107, and 329, respectively. Then, image-description pairs are acquired as outlined in Section 2. The train set is used both for retriever training and as image-modality knowledge incorporated into the corpus.

To ensure the response efficiency, ResNet-18 \cite{21} was conducted as the retriever, which is trained for 500 epochs using the AdamW optimizer with an initial learning rate of $1.0 \times 10^{-4}$ and a cosine learning rate decay strategy. The weight decay is set to $1.0 \times 10^{-4}$, and the batch size is 32. All the slices were normalized using a lung window and resized to 256 × 256 for training, while the corpus is composed of the 369 × 369 resolution. The proposed approach is implemented in PyTorch, and all the methods are conducted on 8 × NVIDIA RTX 4090 GPUs.

\subsection{Qualitative and Quantitative Results}

\begin{table}[t]
	\centering
	\label{table1}
	\caption{Performance comparison of our approach with the comparing methods.}
	\begin{tabularx}{\linewidth}{l>{\centering\arraybackslash}X>{\centering\arraybackslash}X>{\centering\arraybackslash}X>{\centering\arraybackslash}X>{\centering\arraybackslash}X} 
		\toprule
		\textbf{Methods} & Accuracy & Precision & Recall & F1 & Specificity \\
		\midrule 
		ResNet-50-3D \cite{21}&0.5789&0.5714&0.8000&0.6667&0.3333  \\ 
		DenceNet-121-3D \cite{22}&0.6316&0.6154	&0.8000	&0.6957	&0.4444 \\ 
		ResNext-50-3D	\cite{23}&0.6316	&0.6154	&0.8000	&0.6957	&0.4444 \\
		MedicalNet \cite{24}	&0.6316	&0.7143	&0.5000	&0.5882	&0.7778 \\
		M3T \cite{25}&0.6842	&0.7000	&0.7000	&0.7000	&0.6667 \\
		\midrule
            LLaVA-Med \cite{30}	&0.4211	&0.0000	&0.0000	&0.0000	&\textbf{0.8889} \\ 
		GPT-4o \cite{16}&0.5263&0.5294&\textbf{0.9000}&0.6667&0.1111 \\
		\textbf{BHD-RAG + GPT-4o}	&0.6842	&0.7000	&0.7000	&0.7000	&0.6667 \\
		GPT-4-turbo \cite{16}	&0.6316	&0.5882	&0.7407	&0.6316	&0.2222 \\ 
		\textbf{BHD-RAG + GPT-4-turbo}	&\textbf{0.7895}	&\textbf{0.8000}	&0.8000	&\textbf{0.8000}	&0.7778 \\
		\bottomrule
	\end{tabularx}
\end{table}

\begin{figure}[t]
	\includegraphics[width=\textwidth]{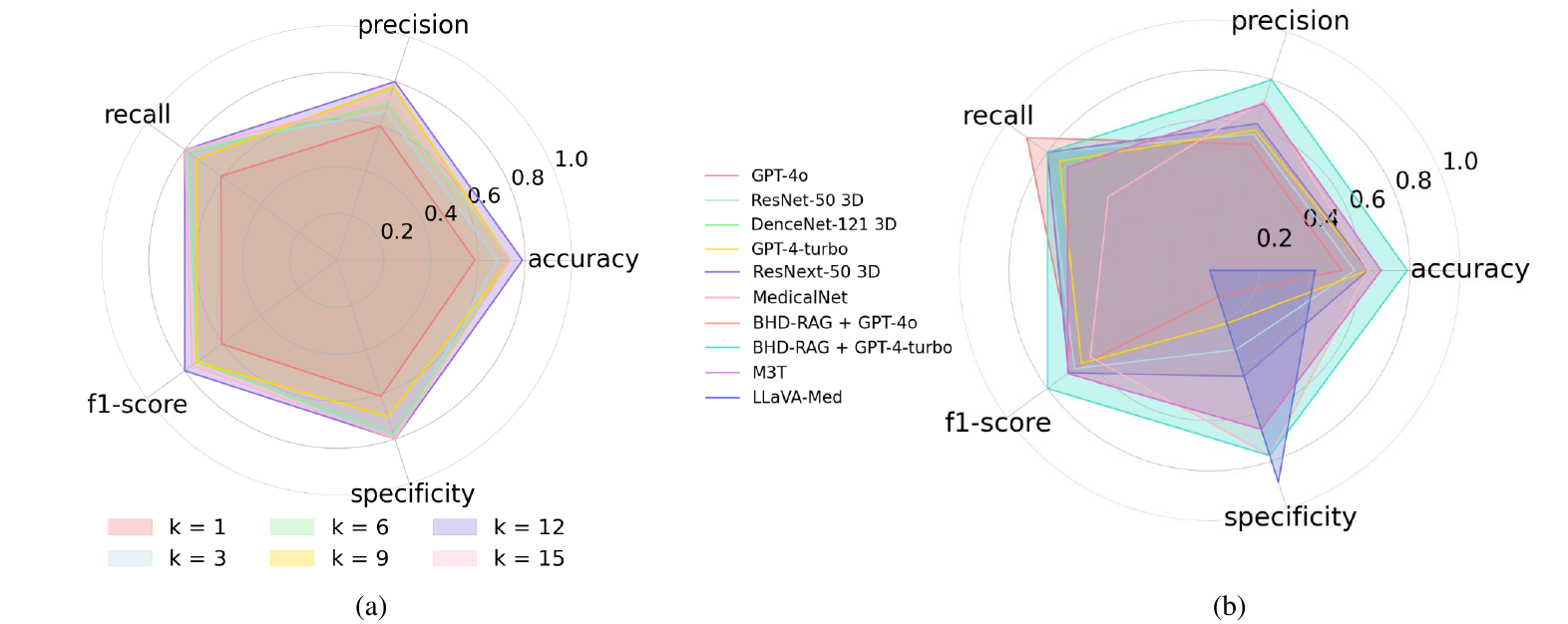}
	\caption{The proposed cosine spatial similarity measure retriever. (a) Influence of $k$ value on BHD-RAG. (b) Quantitative comparison between BHD-RAG and other methods.}
	\label{fig4}
\end{figure}

As shown in Table 1 and Fig. \ref{fig4}, we evaluate the performance of BHD-RAG on the DCLDs dataset using five metrics: accuracy, precision, recall, F1-score, and specificity. Our approach significantly outperforms the best discriminant method, M3T, with metric enhancements ranging from 14.29 \% (precision / recall / F1-score) to 19.99 \% (specificity), while attaining a 15.39 \% accuracy improvement. Additionally, it demonstrates an advantage over MLLMs. BHD-RAG significantly enhances GPT-4o, improving accuracy, precision, and F1-score by 30.00\%, 32.23\%, and 4.99\%, respectively. When integrated with GPT-4-turbo, these gains further increase to 25\%, 36.01\%, and 26.66\%. In contrast, MLLMs without retrieval-augmented generation, particularly GPT-4o, tends to classify all DCLDs samples as BHD, while LLaVA-Med categorizes all DCLDs cases as non-BHD. Although these misclassifications leads to superficially improved recall or Specificity, it lacks clinical significance.

\begin{figure}[t]
	\includegraphics[width=\textwidth]{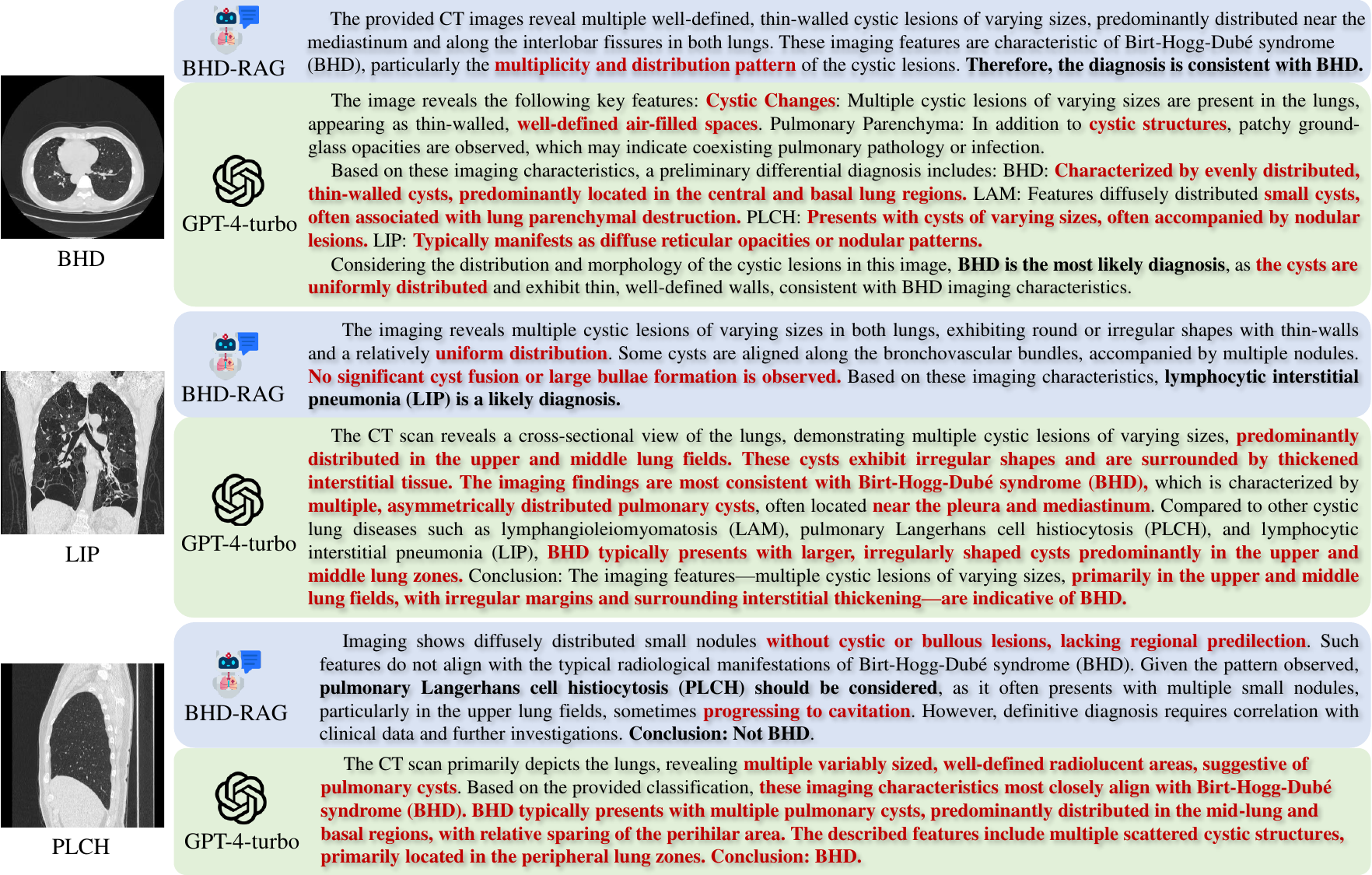}
	\caption{Qualitative comparison of BHD-RAG and GPT-4-turbo diagnoses. Errors or imprecisions in results are highlighted in red by experts.}
	\label{fig5}
\end{figure}

Fig. \ref{fig4} (a) illustrates the impact of $k$ on the performance of BHD-RAG. A larger $k$ enhances stability and $k = 12$ achieves the best performance. Additionally, qualitative comparison of BHD-RAG and GPT-4-turbo are presented in Fig. \ref{fig5}. BHD-RAG enables precise diagnosis of DCLDs and generates evidence-based imaging descriptions, emphasizing subtle features that may be overlooked by experts and MLLMs. In contrast, models without BHD-RAG generate inaccurate, inadequate, hallucination-prone descriptions with colloquial language and insufficient clinical precision.

\begin{table}[tbp]
	\centering
	\label{table2}
	\caption{Ablation experiments of our BHD-RAG on the DCLDs dataset. To observe the performance changes, we remove the retriever and typical features.}
	\begin{tabularx}{\linewidth}{>{\centering\arraybackslash}m{0.6in} >{\centering\arraybackslash}m{0.6in} >{\centering\arraybackslash}m{0.65in} >{\centering\arraybackslash}m{0.65in} >{\centering\arraybackslash}m{0.65in} >{\centering\arraybackslash}m{0.65in} >{\centering\arraybackslash}m{0.65in}}
		\toprule
		Retriever & Typical Features & Accuracy&Precision & Recall & F1 & Specificity \\
		\midrule 
		\usym{2717}&\usym{2717}& 0.6316&	0.5882&	0.7407&	0.6316&	1.0000  \\ 
		\usym{2713}	&\usym{2717}&0.6842&	0.7500	&0.6000&	0.6667&	0.6000 \\ 
		\usym{2717}&\usym{2713}	&0.6842&	0.6667&	0.8000&	0.7273&	0.8000 \\
		\usym{2713}	&\usym{2713}	&0.7895&	0.8000&	0.8000	&0.8000	&0.8000 \\
		\bottomrule
	\end{tabularx}
\end{table}

\subsection{Ablation Study}
We conduct an ablation study on BHD-RAG to verify the effectiveness of the proposed cosine-space similarity retriever and typical features. As shown in Table 2, removing the retriever or typical features reduces accuracy by 13.34 \%, while removing both results in a 20 \% decline. Individually retrieved image-description pairs exhibit limited discriminative features, rendering them suboptimal as external knowledge for DCLDs diagnosis. In contrast, integrating typical differentiating features of DCLDs enhances alignment with clinical diagnoses. This confirms that BHD-RAG enhances the capacity of MLLMs to discern domain-specific DCLDs features through external knowledge integration.

\section{Conclusion}

This study proposes BHD-RAG, a retrieval-augmented generation framework designed for BHD diagnosis. By leveraging a cosine-space similarity retriever, the framework integrates DCLD-specific expertise and clinical precedents with MLLMs to mitigate the inherent hallucination and improve BHD diagnostic accuracy. Evaluation results on the DCLDs dataset demonstrate the effectiveness and generalizability of the proposed method. Future works will focus on expanding the clinical dataset through multi-center collaborations to develop a multimodal foundation model for DCLDs.

%
%
%
\bibliographystyle{splncs04}
\bibliography{mybibliography}

\end{document}